\documentclass[journal]{IEEEtran}
\usepackage{cite}
\usepackage{amsmath,amssymb,amsfonts}
\usepackage{algorithmic}
\usepackage{graphicx}
\usepackage{comment}
\usepackage{multirow}
\usepackage{textcomp}
\usepackage{xcolor}
\usepackage{fancyhdr}
\usepackage{cite}

\usepackage[hidelinks]{hyperref}

\def\BibTeX{{\rm B\kern-.05em{\sc i\kern-.025em b}\kern-.08em
    T\kern-.1667em\lower.7ex\hbox{E}\kern-.125emX}}

\begin{document}

\title{Double-Anonymous Review for Robotics}
\author{Justin K. Yim, Paul Nadan, James Zhu, Alexandra Stutt, J. Joe Payne, Catherine Pavlov, and Aaron M. Johnson%
\thanks{Robomechanics Lab, Carnegie Mellon University,  {\tt\small amj1@cmu.edu}}\\
August 24, 2022}
\maketitle

\section{Introduction}
Prior research has investigated the benefits and costs of double-anonymous review (DAR, also known as double-blind review) in comparison to single-anonymous review (SAR) and open review (OR).
Several review papers have attempted to compile experimental results in peer review research both broadly and in engineering and computer science specifically \cite{snodgrass2006, mckinley2008, Largent2016, Shah2022}.
This document summarizes prior research in peer review that may inform decisions about the format of peer review in the field of robotics and makes some recommendations for potential next steps for robotics publications.

\section{Summary}
Researchers have investigated several possible advantages of DAR:

\begin{itemize}
    \item Improved fairness or reduction in bias:
    \begin{itemize}
        \item based on gender
        \item based on author or institution reputation or prestige
        \item based on nationality
    \end{itemize}
    \item Improved review quality
    \item Reduced perceptions of bias
\end{itemize}

The presence of gender bias and effect of DAR on such bias is a common concern in research into peer review but the conclusions are varied.
Many studies do conclude that gender can disadvantage authors, particularly women \cite{MathildaEffect, wenneras1997} and that DAR can reduce this bias \cite{budden2008}.
However, other studies do not find evidence of gender bias that achieves statistical significance \cite{WEBB2008351, cox2019, Tomkins2017}.
Research into peer review gender bias spans many fields whose different norms and demographics may vary gender bias in field-specific ways.
Notably, gender stereotypes associated with particular sub-domains may bias for or against a particular gender \cite{MathildaEffect}.

Bias in peer review favoring already well-regarded senior authors, institutions, and companies is more conclusively established than gender bias.
Several studies find evidence of this ``status bias" \cite{Sun2021, blank1991, Tomkins2017}, as well as bias in favor of U.S.-based authors \cite{link_1998_US_bias} and against authors that are newcomers to a particular conference \cite{seeber2017}.

Furthermore, \cite{Sun2021} and \cite{laband1994citation} find that this bias may negatively impact the review quality in that double-anonymized reviewers were more effective at separating papers that would achieve high impact from those that would not.

Discussion of potential downsides to DAR largely focuses on difficulty of implementation.
Achieving anonymization is not perfect \cite{bachelli2017, goues_effectiveness_2018} and becoming more difficult due to fast access to digital records including pre-prints \cite{Largent2016}.
In the field of robotics, anonymization may be difficult if particular datasets, robot hardware, test settings, etc are known to be available only to a particular group of authors.
However, even when reviewers self-report as having the highest level of expertise in their field, their guess accuracy is no better than those who are self-reported as less knowledgeable \cite{goues_effectiveness_2018}.
Increased editor burden in handling conflict of interest, author burden in anonymizing the manuscript, and reviewer burden in navigating prior work by others and by the authors are also cited as costs to DAR.

Despite these challenges, numerous robotics conferences have already made the shift to DAR, including RSS and a majority of ACM-sponsored conferences. Furthermore, top machine learning conferences such as NeurIPS and CoRL have implemented both DAR and open review.

Several authors note that DAR reduces perception of bias whether or not it is effective in truly reducing bias \cite{Schulzrinne2009, snodgrass2006}.
Perceptions of bias may have negative consequences on their own if they cause authors to alter their behavior, discourage them from persisting in their field of study, or skew their interpretations of reviewers' comments.


\section{Next Steps}

Based on the current literature, we find that the evidence in support of double-anonymous review is not sufficient to conclusively recommend for implementation in robotics conferences and journals.
One primary drawback that was noted in many of the works reviewed was that they are limited in scope or sample size and do not generalize well to different fields of study.
While many of the studies discussed here were carried out in STEM fields like computer science and medicine, no study of DAR exists in robotics.
Due to this, we recommend organizing boards consider the following action items to better understand what policy is most appropriate to limit bias and inequity in the review process.
\begin{itemize}
    \item Perform a study of one or more major robotics conferences in which half or all papers are double-anonymized to capture any differences between SAR and DAR policies. A full conference study would be the easiest, but splitting a conference at the reviewer, paper, or associate editor level would provide for a better comparison. This would not require doubling the number of reviews, as done in \cite{lawrence_neurips_retrospective}.
    \item Survey the contributors and reviewers of the conference or journal to gauge support of DAR and perceptions of bias in the current system.
    \item Change policies other than DAR, such as improving representation among the senior editors as recommended in \cite{Schulzrinne2009} or revising reviewer training and prompts.
\end{itemize}

\section{Literature Review}


The following section summarizes each work referenced above, organized by topic: gender bias, status bias, paper quality, implementation challenges and perception, and reviews and summaries.

\subsection{Gender Bias}

\cite{MathildaEffect} found that research attributed to male authors is viewed as more scientifically robust than identical research attributed to female authors, for both male and female reviewers. This effect was particularly strong in fields typically associated with ``masculine" traits. Additionally, the interest in collaboration with authors was higher for work in which the author's gender matched that of the gender stereotype for their research topic.

\cite{budden2008} compared gender representation in two similar behavioral ecology journals, Behavioral Ecology (BE), and Behavioral Ecology and Sociobiology (BES). BE implemented double-anonymous review in 2001, and from 2002-2005 BE saw a $7.9\%$ increase in the proportion of female first-authored papers, while BES saw no significant change. The number of papers published by each journal has increased since 1997, and the authors found that the mean number of citations per paper was comparable in both journals, leading them to conclude that the switch from single-anonymous to double-anonymous review had no adverse effect on the quantity or impact of publications.
However, \cite{WEBB2008351} re-analyzed the data with a different method and found that BE's increase in female first-authored papers was not significantly different from other journals in the field.

In \cite{cox2019}, a review of behavioral ecology journals found no evidence of double-anonymous review affecting gender bias. Their data set included one double-anonymous and four single-anonymous journals, for a total of 4,865 papers. The authors discuss the administrative costs of double-anonymous review (burden on authors and editors, conflicts of interest, preventing use of preprint servers, and prestige bias toward more easily identifiable authors) compared to potential benefits (preventing nepotism and institutional/geographic biases). However, this study makes no effort to control for other differences between journals (e.g. by comparing both review processes for submissions to the same journal). Further considering that only one double-anonymous review journal was considered in the sample, the results of this study are not compelling.

\subsection{Status Bias}

The tendency to over-recognize the contributions of already well-regarded authors was noted and termed the ``Matthew Effect" in \cite{MatthewEffect}.

\cite{Sun2021} found that when the Conference on Learning Representations (ICLR) moved from single-anonymous to double-anonymous review, the scores given to the most prestigious authors decreased significantly.
They also found that double-anonymous review was more effective at accepting high quality papers (those that are cited most often) and rejecting low quality papers (those cited least often).

In \cite{blank1991}, the author examines status bias by randomly assigning submissions to The American Economic Review to either SAR or DAR. They find that the acceptance rates for DAR vs SAR papers are similar for both top (rank 1-5) and low-ranked (rank $>$50) institutions, but decreased by 7.5\% and 4.9\% for near top (rank 6-20) and mid-range (rank 21-50) institutions respectively in their sample of 1,498 papers. U.S. research institutions also see a 5.9\% decrease. However, all of these results are only marginally significant. The author also finds that 48.6\% of reviewers were able to correctly identify the authors of the papers, and observe a small but not statistically significant effect (due to the small sample size of female authors) of DAR in reducing gender bias.

The ACM Web Search and Data Mining (WSDM) 2017 conference performed a controlled experiment comparing single- and double-anonymous reviews \cite{Tomkins2017}. Regarding review recommendations, they investigate three main hypotheses: 1) papers by female authors are viewed less favorably, 2) papers by famous authors are viewed more favorably, and 3) papers from famous institutions and companies are viewed more favorably, as well as three other factors (American papers, reviewer and authors from the same country, and academic institutions vs. corporate/government). They found that reviewers were significantly more likely to recommend papers for acceptance from famous authors, top institutions, and top companies by factors of 1.63, 1.58, and 2.10 respectively when given access to authorship information. They did not find significant results based on gender despite trying multiple different formulations. The study concludes that single-anonymous reviewers are using author and institution information (for better or worse) and that organizers should consider the advantages of double-anonymous reviewing. There are several limitations to this study: only reviews and not final acceptance decisions are addressed and it is possible that reviewer bidding differences in the single- and double- anonymized cases may have subtly altered the review assignments resulting in some difference in the paper scoring.

In \cite{seeber2017}, the authors study 21,535 papers from 71 CS conferences to determine the relative share published by researchers who are new to that conference. They find that newcomers are around twice as common in conferences that use double anonymous review than single, with the effect strongest for more experienced researchers and nonexistent for newcomers to the field as a whole (based on total number of papers published in the field). From this result, the authors conclude that single anonymous review leads to reviewer bias against researchers who are not already in the reviewers' research community, which can negatively impact innovation and creativity.

\subsection{Paper Quality}

The 2014 NeurIPS conference was studied for consistency of reviews (using double-anonymous review) \cite{lawrence_neurips_retrospective}. A random sample of submissions were reviewed a second time by a different group of reviewers. It was found that there was significant inconsistency in the review committees, (an expected proportion of between $38\%$ and $64\%$ of accepted papers to be the same), but it is better than random (expected $25\%$ of same papers).

\cite{laband1994citation} compared the effectiveness of separating papers that would have higher impact (as measured by citations) from those that would have lower impact among economics journals using SAR and DAR.
1051 articles published in 15 SAR and 13 DAR journals form the dataset.
Controlling for article length and journal quality, SAR journals show a higher rate of ``type I error" in which lower impact papers are erroneously accepted.
However, this is an observational study and not an experimental study and correlations between journal quality and review style may make the significance of these results less clear.

\subsection{Implementation Challenges and Public Perception}

\cite{goues_effectiveness_2018} examined how well reviewers were able to guess the identity of an author in a double-anonymous review.
They found that over 3 different software engineering conferences in 2016, the majority of papers ($>74\%$) did not have any correct author guesses, and a small amount ($<7\%$) had all author guesses correct.
This work reveals that while double-anonymous review is not perfect, it does an effective job in masking the identity of the authors.

In \cite{bachelli2017} the authors investigate the costs and benefits of double-anonymous review in the context of software engineering conferences. They come to the conclusion that the costs of double-anonymous review outweigh the benefits. However, in this paper, all of the results are based on surveys with members of the community rather than outcome based results. This is highly limited especially since many biases are subconscious rather than explicit. People may also overstate the costs of double-anonymous review if they are opposed to implementation of it.

In \cite{Schulzrinne2009}, the author argues that the implementation of double-anonymous review does little to actually improve the fairness of the review process and instead acts as ``review theater", like the ``security theater" of the TSA. They argue that it is often difficult or impossible to truly obscure author identities through the complete review process. However, the author does recognize the importance of the perception of fairness, which double anonymous review may aid in. Ultimately, the author views the implementation of double anonymous review as something that is often performed in place of more difficult changes to the review process such as ensuring that the reviewers and the editorial committee appropriately reflect the composition of those submitting to the conferences.

\subsection{Reviews and Summaries}
In 2007 the Association for Computing Machinery (ACM) Transactions on Database Systems (TODS) adopted double-anonymous reviews after discussion with the community and concerns about fairness of single-anonymous reviews \cite{snodgrass2007}. Snodgrass, as editor-in-chief of this journal, wrote a literature review comparing SAR and DAR prior to the journal's decision \cite{snodgrass2006}. Snodgrass finds evidence of status bias for mid-range institutions and U.S. based authors, although the evidence is mixed for top-range institutions. They also find evidence of gender bias, and use the MLA journal as an example of a switch to DAR in 1974 that led to a dramatic increase in accepted submissions from female authors. They then enumerate and address many proposed counterarguments to DAR, including author/reviewer burden, conflicts of interest, and dissemination of preprints. The author concludes with an anecdote showing that support for DAR is mixed among the scholarly elite, but overwhelming among the general population of qualified authors.

In \cite{mckinley2008} the author presents a summary of a variety of studies to support double-anonymous peer review \cite{laband1994citation,wenneras1997,watson2005mysterious}.
This includes studies demonstrating that women are significantly less likely to be accepted for fellowships and post-doctoral positions; in one instance, women needed to be 2.5 times as productive as men to be judged to be as good as them. 
This study also found that even if committee members with conflicts of interest were prevented from reviewing a candidate, other committee members would still score those candidates higher, the extent to which the scores were inflated was approximately equivalent to the candidate having 3 more Nature or Science articles.
Another source cited showed that in one area, papers accepted through double anonymous review had higher citation rates than those accepted through single anonymous.
The author then works through a variety of difficulties with double-anonymous review, including handling conflicts in software, de-anonymizing authors to avoid missed conflicts, and creating an external review committee.
Lastly, the author argues in favor of an author response phase during the review process.

\cite{Largent2016} is a review of research into double-anonymous review.
Their review shows inconclusive evidence for effects on review bias, possibly no relationship for review quality (and similarly mixed results for open review's effect on review quality), mixed effectiveness in achieving anonymization (and increasing ease of breaking anonymization online), but widespread support for anonymization and perception of bias in single-anonymous review.
The review concludes that mere perception of bias can be reason to implement double-anonymous review but recommend that each publication make a careful decision on a case-by-case basis.

\bibliography{References}
\bibliographystyle{ieeetr}
\end{document}